# Efficient Minimization of Decomposable Submodular Functions

**Peter Stobbe**
California Institute of Technology
Pasadena, CA 91125
stobbe@caltech.edu

**Andreas Krause**
California Institute of Technology
Pasadena, CA 91125
krausea@caltech.edu

## Abstract

Many combinatorial problems arising in machine learning can be reduced to the problem of minimizing a submodular function. Submodular functions are a natural discrete analog of convex functions, and can be minimized in strongly polynomial time. Unfortunately, state-of-the-art algorithms for general submodular minimization are intractable for larger problems. In this paper, we introduce a novel subclass of submodular minimization problems that we call *decomposable*. Decomposable submodular functions are those that can be represented as sums of concave functions applied to modular functions. We develop an algorithm, SLG, that can efficiently minimize decomposable submodular functions with tens of thousands of variables. Our algorithm exploits recent results in smoothed convex minimization. We apply SLG to synthetic benchmarks and a joint classification-and-segmentation task, and show that it outperforms the state-of-the-art general purpose submodular minimization algorithms by several orders of magnitude.

## 1 Introduction

Convex optimization has become a key tool in many machine learning algorithms. Many seemingly multimodal optimization problems such as nonlinear classification, clustering and dimensionality reduction can be cast as convex programs. When minimizing a convex loss function, we can rest assured to efficiently find an optimal solution, even for large problems. Convex optimization is a structural property of continuous optimization problems. However, many machine learning problems, such as structure learning, variable selection, MAP inference in discrete graphical models, require solving discrete, combinatorial optimization problems.

In recent years, another fundamental problem structure, which has similar beneficial properties, has emerged as very useful in many combinatorial optimization problems arising in machine learning: Submodularity is an intuitive diminishing returns property, stating that adding an element to a smaller set helps more than adding it to a larger set. Similarly to convexity, submodularity allows one to efficiently find provably (near-)optimal solutions. In particular, the minimum of a submodular function can be found in strongly polynomial time [11]. Unfortunately, while polynomial-time solvable, exact techniques for submodular minimization require a number of function evaluations on the order of $n^5$ [12], where $n$ is the number of variables in the problem (e.g., number of random variables in the MAP inference task), rendering the algorithms impractical for many real-world problems.

Fortunately, several submodular minimization problems arising in machine learning have structure that allows solving them more efficiently. Examples include symmetric functions that can be solved in $O(n^3)$ evaluations using Queyranne's algorithm [19], and functions that decompose into attractive, pairwise potentials, that can be solved using graph cutting techniques [7]. In this paper, we introduce a novel class of submodular minimization problems that can be solved efficiently. In particular, we develop an algorithm SLG, that can minimize a class of submodular functions that we call *decomposable*: These are functions that can be decomposed into sums of concave functions applied to modular (additive) functions. Our algorithm is based on recent techniques of smoothed convex minimization [18] applied to the Lovász extension. We demonstrate the usefulness of our algorithm on a joint classification-and-segmentation task involving tens of thousands of variables, and show that it outperforms state-of-the-art algorithms for general submodular function minimization by several orders of magnitude.

## 2 Background on Submodular Function Minimization

We are interested in minimizing set functions that map subsets of some base set $E$ to real numbers. I.e., given $f : 2^E \to \mathbb{R}$ we wish to solve for $A^* \in \arg\min_A f(A)$. For simplicity of notation, we use the base set $E = \{1, \ldots n\}$, but in an application the base set may consist of nodes of a graph, pixels of an image, etc. Without loss of generality, we assume $f(\emptyset) = 0$. If the function $f$ has no structure, then there is no way solve the problem other than checking all $2^n$ subsets. In this paper, we consider functions that satisfy a key property that arises in many applications: *submodularity* (c.f., [16]). A set function $f$ is called submodular iff, for all $A, B \in 2^E$, we have

$$f(A \cup B) + f(A \cap B) \leq f(A) + f(B). \tag{1}$$

Submodular functions can alternatively, and perhaps more intuitively, be characterized in terms of their discrete derivatives. First, we define $\Delta_k f(A) = f(A \cup \{k\}) - f(A)$ to be the discrete derivative of $f$ with respect to $k \in E$ at $A$; intuitively this is the change in $f$'s value by adding the element $k$ to the set $A$. Then, $f$ is submodular iff:

$$\Delta_k f(A) \geq \Delta_k f(B), \text{ for all } A \subseteq B \subseteq E \text{ and } k \in E \setminus B.$$

Note the analogy to concave functions; the discrete derivative is smaller for larger sets, in the same way that $\phi(x+h) - \phi(x) \geq \phi(y+h) - \phi(y)$ for all $x \leq y$, $h \geq 0$ if and only if $\phi$ is a concave function on $\mathbb{R}$. Thus a simple example of a submodular function is $f(A) = \phi(|A|)$ where $\phi$ is any concave function. Yet despite this connection to concavity, it is in fact 'easier' to minimize a submodular function than to maximize it[1], just as it is easier to minimize a convex function. One explanation for this is that submodular minimization can be reformulated as a convex minimization problem.

To see this, consider taking a set function minimization problem, and reformulating it as a minimization problem over the unit cube $[0, 1]^n \subset \mathbb{R}^n$. Define $\boldsymbol{e}_A \in \mathbb{R}^n$ to be the indicator vector of the set $A$, i.e.,

$$\boldsymbol{e}_A[k] = \begin{cases} 0 \text{ if } k \notin A \\ 1 \text{ if } k \in A \end{cases}$$

We use the notation $\boldsymbol{x}[k]$ for the $k$th element of the vector $\boldsymbol{x}$. Also we drop brackets and commas in subscripts, so $\boldsymbol{e}_{kl} = \boldsymbol{e}_{\{k,l\}}$ and $\boldsymbol{e}_k = \boldsymbol{e}_{\{k\}}$ as with the standard unit vectors. A continuous extension of a set function $f$ is a function $\tilde{f}$ on the unit cube $\tilde{f} : [0,1]^n \to \mathbb{R}$ with the property that $f(A) = \tilde{f}(\boldsymbol{e}_A)$. In order to be useful, however, one needs the minima of the set function to be related to minima of the extension:

$$A^* \in \arg\min_{A \in 2^E} f(A) \Rightarrow \boldsymbol{e}_{A^*} \in \arg\min_{\boldsymbol{x} \in [0,1]^n} \tilde{f}(\boldsymbol{x}). \tag{2}$$

A key result due to Lovász [16] states that each submodular function $f$ has an extension $\tilde{f}$ that not only satisfies the above property, but is also convex and efficient to evaluate. We can define the *Lovász extension* in terms of the submodular polyhedron $P_f$:

$$P_f = \{\boldsymbol{v} \in \mathbb{R}^n : \boldsymbol{v} \cdot \boldsymbol{e}_A \leq f(A), \text{ for all } A \in 2^E\}, \quad \tilde{f}(\boldsymbol{x}) = \sup_{v \in P_f} \boldsymbol{v} \cdot \boldsymbol{x}.$$

The submodular polyhedron $P_f$ is defined by exponentially many inequalities, and evaluating $\tilde{f}$ requires solving a linear program over this polyhedron. Perhaps surprisingly, as shown by Lovász, $\tilde{f}$ can be very efficiently computed as follows. For a fixed $\boldsymbol{x}$ let $\sigma : E \to E$ be a permutation such that $\boldsymbol{x}[\sigma(1)] \geq \ldots \geq \boldsymbol{x}[\sigma(n)]$, and then define the set $S_k = \{\sigma(1), \ldots, \sigma(k)\}$. Then we have a formula for $\tilde{f}$ and a subgradient:

$$\tilde{f}(\boldsymbol{x}) = \sum_{k=1}^n \boldsymbol{x}[\sigma(k)](f(S_k) - f(S_{k-1})), \quad \partial \tilde{f}(\boldsymbol{x}) \ni \sum_{k=1}^n \boldsymbol{e}_{\sigma(k)}(f(S_k) - f(S_{k-1})).$$

Note that if two components of $\boldsymbol{x}$ are equal, the above formula for $\tilde{f}$ is independent of the permutation chosen, but the subgradient is not unique.

---

[1] With the additional assumption that $f$ is nondecreasing, *maximizing* a submodular function subject to a cardinality constraint $|A| \leq M$ is 'easy'; a greedy algorithm is known to give a near-optimal answer [17].

Equation (2) was used to show that submodular minimization can be achieved in polynomial time [16]. However, algorithms which directly minimize the Lovasz extension are regarded as impractical. Despite being convex, the Lovász extension is non-smooth, and hence a simple subgradient descent algorithm would need $O(1/\epsilon^2)$ steps to achieve $O(\epsilon)$ accuracy.

Recently, Nesterov showed that if knowledge about the structure of a particular non-smooth convex function is available, it can be exploited to achieve a running time of $O(1/\epsilon)$ [18]. One way this is done is to construct a smooth approximation of the non-smooth function, and then use an accelerated gradient descent algorithm which is highly effective for smooth functions. Connections of this work with submodularity and combinatorial optimization are also explored in [4] and [2]. In fact, in [2], Bach shows that computing the smoothed Lovász gradient of a general submodular function is equivalent to solving a submodular minimization problem. In this paper, we do not treat general submodular functions, but rather a large class of submodular minimization functions that we call *decomposable*. (To apply the smoothing technique of [18], special structural knowledge about the convex function is required, so it is natural that we would need special structural knowledge about the submodular function to leverage those results.) We further show that we can exploit the discrete structure of submodular minimization in a way that allows terminating the algorithm early with a certificate of optimality, which leads to drastic performance improvements.

## 3  The Decomposable Submodular Minimization Problem

In this paper, we consider the problem of minimizing functions of the following form:

$$f(A) = \boldsymbol{c} \cdot \boldsymbol{e}_A + \sum_j \phi_j(\boldsymbol{w}_j \cdot \boldsymbol{e}_A), \tag{3}$$

where $\boldsymbol{c}, \boldsymbol{w}_j \in \mathbb{R}^n$ and $\boldsymbol{0} \leq \boldsymbol{w}_j \leq \boldsymbol{1}$ and $\phi_j : [0, \boldsymbol{w}_j \cdot \boldsymbol{1}] \to \mathbb{R}$ are arbitrary concave functions. It is shown in the Appendix that functions of this form are submodular. We call this class of functions *decomposable submodular functions*, as they decompose into a sum of concave functions applied to nonnegative modular functions[2]. Below, we give examples of decomposable submodular functions arising in applications.

We first focus on the special case where all the concave functions are of the form $\phi_j(\cdot) = d_j \min(y_j, \cdot)$ for some $y_j, d_j > 0$. Since these potentials are of key importance, we define the submodular functions $\Psi_{\boldsymbol{w},y}(A) = \min(y, \boldsymbol{w} \cdot \boldsymbol{e}_A)$ and call them *threshold potentials*. In Section 5, we will show in how to generalize our approach to arbitrary decomposable submodular functions.

**Examples.** The simplest example is a *2-potential*, which has the form $\phi(|A \cap \{k, l\}|)$, where $\phi(1) - \phi(0) \geq \phi(1) - \phi(2)$. It can be expressed as a sum of a modular function and a threshold potential:

$$\phi(|A \cap \{k, l\}|) = \phi(0) + (\phi(2) - \phi(1))\boldsymbol{e}_{kl} \cdot \boldsymbol{e}_A + (2\phi(1) - \phi(0) - \phi(2))\Psi_{\boldsymbol{e}_{kl},1}(A)$$

Why are such potential functions interesting? They arise, for example, when finding the Maximum a Posteriori configuration of a pairwise Markov Random Field model in image classification schemes such as in [20]. On a high level, such an algorithm computes a value $\boldsymbol{c}[k]$ that corresponds to the log-likelihood of pixel $k$ being of one class vs. another, and for each pair of adjacent pixels, a value $d_{kl}$ related to the log-likelihood that pixels $k$ and $l$ are of the same class. Then the algorithm classifies pixels by minimizing a sum of 2-potentials: $f(A) = \boldsymbol{c} \cdot \boldsymbol{e}_A + \sum_{k,l} d_{kl}(1 - |1 - \boldsymbol{e}_{kl} \cdot \boldsymbol{e}_A|)$. If the value $d_{kl}$ is large, this encourages the pixels $k$ and $l$ to be classified similarly.

More generally, consider a higher order potential function: a concave function of the number of elements in some activation set $S$, $\phi(|A \cap S|)$ where $\phi$ is concave. It can be shown that this can be written as a sum of a modular function and a positive linear combination of $|S| - 1$ threshold potentials. Recent work [14] has shown that classification performance can be improved by adding terms corresponding to such higher order potentials $\phi_j(|R_j \cap A|)$ to the objective function where the functions $\phi_j$ are piecewise linear concave functions, and the regions $R_j$ of various sizes generated from a segmentation algorithm. Minimization of these particular potential functions can then be reformulated as a graph cut problem [13], but this is less general than our approach.

Another canonical example of a submodular function is a set cover function. Such a function can be reformulated as a combination of concave cardinality functions (details in appendix). So all functions which are weighted combinations of set cover functions can be expressed as threshold potentials. However, threshold potentials with nonuniform weights are strictly more general than concave cardinality potentials. That is, there exists $w$ and $y$ such that $\Psi_{w,y}(A)$ cannot be expressed as $\sum_j \phi_j(|R_j \cap A|)$ for *any* collection of concave $\phi_j$ and sets $R_j$.

---
[2]A function is called *modular* if (1) holds with equality. It can be written as $A \mapsto \boldsymbol{w} \cdot \boldsymbol{e}_A$ for some $\boldsymbol{w} \in \mathbb{R}^n$.

Another example of decomposable functions arises in multiclass queuing systems [10]. These are of the form $f(A) = c \cdot e_A + u \cdot e_A \phi(v \cdot e_A)$, where $u, v$ are nonnegative weight vectors and $\phi$ is a nonpositive nonincreasing concave function. With the proper choice of $\phi_j$ and $w_j$ (again details are in appendix), this can in fact be reformulated as sum of the type in Eq. 3 with $n$ terms.

In our own experiments, shown in Section 6, we use an implementation of TextonBoost [20] and augment it with quadratic higher order potentials. That is, we use TextonBoost to generate per-pixel scores $c$, and then minimize $f(A) = c \cdot e_A + \sum_j |A \cap R_j||R_j \setminus A|$, where the regions $R_j$ are regions of pixels that we expect to be of the same class (e.g., by running a cheap region-growing heuristic). The potential function $|A \cap R_j||R_j \setminus A|$ is smallest when $A$ contains all of $R_j$ or none of it. It gives the largest penalty when exactly half of $R_j$ is contained in $A$. This encourages the classification scheme to classify most of the pixels in a region $R_j$ the same way. We generate regions with a basic region-growing algorithm with random seeds. See Figure 1(a) for an illustration of examples of regions that we use. In our experience, this simple idea of using higher-order potentials can dramatically increase the quality of the classification over one using only 2-potentials, as can be seen in Figure 2.

## 4 The SLG Algorithm for Threshold Potentials

We now present our algorithm for efficient minimization of a decomposable submodular function $f$ based on smoothed convex minimization. We first show how we can efficiently smooth the Lovász extension of $f$. We then apply accelerated gradient descent to the gradient of the smoothed function. Lastly, we demonstrate how we can often obtain a certificate of optimality that allows us to stop early, drastically speeding up the algorithm in practice.

### 4.1 The Smoothed Extension of a Threshold Potential

The key challenge in our algorithm is to efficiently smooth the Lovász extension of $f$, so that we can resort to algorithms for accelerated convex minimization. We now show how we can efficiently smooth the *threshold potentials* $\Psi_{w,y}(A) = \min(y, w \cdot e_A)$ of Section 3, which are simple enough to allow efficient smoothing, but rich enough when combined to express a large class of submodular functions. For $x \geq 0$, the Lovász extension of $\Psi_{w,y}$ is

$$\tilde{\Psi}_{w,y}(x) = \sup v \cdot x \text{ s.t. } v \leq w, v \cdot e_A \leq y \text{ for all } A \in 2^E.$$

Note that when $x \geq 0$, the arg max of the above linear program always contains a point $v$ which satisfies $v \cdot \mathbf{1} = y$, and $v \geq 0$. So we can restrict the domain of the dual variable $v$ to those points which satisfy these two conditions, without changing the value of $\tilde{\Psi}(x)$:

$$\tilde{\Psi}_{w,y}(x) = \max_{v \in \mathcal{D}(w,y)} v \cdot x \text{ where } \mathcal{D}(w,y) = \{v : \mathbf{0} \leq v \leq w, v \cdot \mathbf{1} = y\}.$$

Restricting the domain of $v$ allows us to define a smoothed Lovász extension (with parameter $\mu$) that is easily computed:

$$\tilde{\Psi}^\mu_{w,y}(x) = \max_{v \in \mathcal{D}(w,y)} v \cdot x - \frac{\mu}{2}\|v\|^2$$

To compute the value of this function we need to solve for the optimal vector $v^*$, which is also the gradient of this function, as we have the following characterization:

$$\nabla \tilde{\Psi}^\mu_{w,y}(x) = \arg\max_{v \in \mathcal{D}(w,y)} v \cdot x - \frac{\mu}{2}\|v\|^2 = \arg\min_{v \in \mathcal{D}(w,y)} \left\|\frac{x}{\mu} - v\right\|. \tag{4}$$

To derive an expression for $v^*$, we begin by forming the Lagrangian and deriving the dual problem:

$$\tilde{\Psi}^\mu_{w,y}(x) = \min_{t \in \mathbb{R}, \lambda_1, \lambda_2 \geq \mathbf{0}} \left(\max_{v \in \mathbb{R}^n} v \cdot x - \frac{\mu}{2}\|v\|^2 + \lambda_1 \cdot v + \lambda_2 \cdot (w - v) + t(y - v \cdot \mathbf{1})\right)$$

$$= \min_{t \in \mathbb{R}, \lambda_1, \lambda_2 \geq \mathbf{0}} \frac{1}{2\mu}\|x - t\mathbf{1} + \lambda_1 - \lambda_2\|^2 + \lambda_2 \cdot w + ty.$$

If we fix $t$, we can solve for the optimal dual variables $\lambda_1^*$ and $\lambda_2^*$ componentwise. By strong duality, we know the optimal primal variable is given by $v^* = \frac{1}{\mu}(x - t^*\mathbf{1} + \lambda_1^* - \lambda_2^*)$. So we have:

$$\lambda_1^* = \max(t^*\mathbf{1} - x, \mathbf{0}), \; \lambda_2^* = \max(x - t^*\mathbf{1} - \mu w, \mathbf{0}) \Rightarrow v^* = \min\left(\max\left((x - t^*\mathbf{1})/\mu, \mathbf{0}\right), w\right).$$

This expresses $\boldsymbol{v}^*$ as a function of the unknown optimal dual variable $t^*$. For the simple case of 2-potentials, we can solve for $t^*$ explicitly and get a closed form expression:

$$\nabla \tilde{\Psi}^\mu_{\boldsymbol{e}_{kl},1}(\boldsymbol{x}) = \begin{cases} \boldsymbol{e}_k & \text{if } \boldsymbol{x}[k] \geq \boldsymbol{x}[l] + \mu \\ \boldsymbol{e}_l & \text{if } \boldsymbol{x}[l] \geq \boldsymbol{x}[k] + \mu \\ \frac{1}{2}(\boldsymbol{e}_{kl} + \frac{1}{\mu}(\boldsymbol{x}[k] - \boldsymbol{x}[l])(\boldsymbol{e}_k - \boldsymbol{e}_l)) & \text{if } |\boldsymbol{x}[k] - \boldsymbol{x}[l]| < \mu \end{cases}$$

However, in general to find $t^*$ we note that $\boldsymbol{v}^*$ must satisfy $\boldsymbol{v}^* \cdot \mathbf{1} = y$. So define $\rho^\mu_{\boldsymbol{x},\boldsymbol{w}}(t)$ as:

$$\rho^\mu_{\boldsymbol{x},\boldsymbol{w}}(t) = \min(\max((\boldsymbol{x} - t\mathbf{1})/\mu, \mathbf{0}), \boldsymbol{w}) \cdot \mathbf{1}$$

Then we note this function is a monotonic continuous piecewise linear function of $t$, so we can use a simple root-finding algorithm to solve $\rho^\mu_{\boldsymbol{x},\boldsymbol{w}}(t^*) = y$. This root finding procedure will take no more than $O(n)$ steps in the worst case.

### 4.2 The SLG Algorithm for Minimizing Sums of Threshold Potentials

Stepping beyond a single threshold potential, we now assume that the submodular function to be minimized can be written as a nonnegative linear combination of threshold potentials and a modular function, i.e.,

$$f(A) = \boldsymbol{c} \cdot \boldsymbol{e}_A + \sum_j d_j \Psi_{\boldsymbol{w}_j, y_j}(A).$$

Thus, we have the smoothed Lovász extension, and its gradient:

$$\tilde{f}^\mu(\boldsymbol{x}) = \boldsymbol{c} \cdot \boldsymbol{x} + \sum_j d_j \tilde{\Psi}^\mu_{\boldsymbol{w}_j, y_j}(\boldsymbol{x}) \text{ and } \nabla \tilde{f}^\mu(\boldsymbol{x}) = \boldsymbol{c} + \sum_j d_j \nabla \tilde{\Psi}^\mu_{\boldsymbol{w}_j, y_j}(\boldsymbol{x}).$$

We now wish to use the accelerated gradient descent algorithm of [18] to minimize this function. This algorithm requires that the smoothed objective has a Lipschitz continuous gradient. That is, for some constant $L$, it must hold that $\|\nabla \tilde{f}^\mu(\boldsymbol{x}_1) - \nabla \tilde{f}^\mu(\boldsymbol{x}_2)\| \leq L\|\boldsymbol{x}_1 - \boldsymbol{x}_2\|$, for all $\boldsymbol{x}_1, \boldsymbol{x}_2 \in \mathbb{R}^n$. Fortunately, by construction, the smoothed threshold extensions $\tilde{\Psi}^\mu_{\boldsymbol{w}_j, y_j}(\boldsymbol{x})$ all have $1/\mu$ Lipschitz gradient, a direct consequence of the characterization in Equation 4. Hence we have a loose upper bound for the Lipschitz constant of $\tilde{f}^\mu$: $L \leq \frac{D}{\mu}$, where $D = \sum_j d_j$. Furthermore, the smoothed threshold extensions approximate the threshold extensions uniformly: $|\tilde{\Psi}^\mu_{\boldsymbol{w}_j, y_j}(\boldsymbol{x}) - \tilde{\Psi}_{\boldsymbol{w}_j, y_j}(\boldsymbol{x})| \leq \frac{\mu}{2}$ for all $\boldsymbol{x}$, so $|\tilde{f}^\mu(\boldsymbol{x}) - \tilde{f}(\boldsymbol{x})| \leq \frac{\mu D}{2}$.

One way to use the smoothed gradient is to specify an accuracy $\varepsilon$, then minimize $\tilde{f}^\mu$ for sufficiently small $\mu$ to guarantee that the solution will also be an approximate minimizer of $\tilde{f}$. Then we simply apply the accelerated gradient descent algorithm of [18]. See also [3] for a description. Let $P_C(\boldsymbol{x}) = \arg\min_{\boldsymbol{x}' \in C} \|\boldsymbol{x} - \boldsymbol{x}'\|$ be the projection of $\boldsymbol{x}$ onto the convex set $C$. In particular, $P_{[0,1]^n}(\boldsymbol{x}) = \min(\max(\boldsymbol{x}, \mathbf{0}), \mathbf{1})$. Algorithm 1 formalizes our *Smoothed Lovász Gradient (*SLG*)* algorithm:

---
**Algorithm 1:** SLG: Smoothed Lovász Gradient

**Input**: Accuracy $\varepsilon$; decomposable function $f$.
**begin**
    $\mu = \frac{\varepsilon}{2D}$, $L = \frac{D}{\mu}$, $\boldsymbol{x}_{-1} = \boldsymbol{z}_{-1} = \frac{1}{2}\mathbf{1}$;
    **for** $t = 0, 1, 2, \ldots$ **do**
        $\boldsymbol{g}_t = \nabla \tilde{f}^\mu(\boldsymbol{x}_{t-1})/L$; $\boldsymbol{z}_t = P_{[0,1]^n}\left(\boldsymbol{z}_{-1} - \sum_{s=0}^t \left(\frac{s+1}{2}\right)\boldsymbol{g}_s\right)$; $\boldsymbol{y}_t = P_{[0,1]^n}(\boldsymbol{x}_t - \boldsymbol{g}_t)$;
        **if** $gap_t \leq \varepsilon/2$ **then stop**;
        $\boldsymbol{x}_t = (2\boldsymbol{z}_t + (t+1)\boldsymbol{y}_t)/(t+3)$;
    $\boldsymbol{x}_\varepsilon = \boldsymbol{y}_t$;
**Output**: $\varepsilon$-optimal $\boldsymbol{x}_\varepsilon$ to $\min_{\boldsymbol{x} \in [0,1]^n} \tilde{f}(\boldsymbol{x})$

---

The optimality gap of a smooth convex function at the iterate $\boldsymbol{y}_t$ can be computed from its gradient:

$$\text{gap}_t = \max_{\boldsymbol{x} \in [0,1]^n}(\boldsymbol{y}_t - \boldsymbol{x}) \cdot \nabla \tilde{f}^\mu(\boldsymbol{y}_t) = \boldsymbol{y}_t \cdot \nabla \tilde{f}^\mu(\boldsymbol{y}_t) + \max(-\nabla \tilde{f}^\mu(\boldsymbol{y}_t), \mathbf{0}) \cdot \mathbf{1}.$$

In summary, as a consequence of the results of [18], we have the following guarantee about SLG:

**Theorem 1** SLG *is guaranteed to provide an $\varepsilon$-optimal solution after running for $\mathcal{O}(\frac{D}{\varepsilon})$ iterations.*

SLG is only guaranteed to provide an $\varepsilon$-optimal solution to the *continuous* optimization problem. Fortunately, once we have an $\varepsilon$-optimal point for the Lovász extension, we can efficiently round it to set which is $\varepsilon$-optimal for the original submodular function using Alg. 2 (see [9] for more details).

---

**Algorithm 2:** Set generation by rounding the continuous solution

**Input**: Vector $\boldsymbol{x} \in [0,1]^n$; submodular function $f$.
**begin**
  By sorting, find any permutation $\sigma$ satisfying: $\boldsymbol{x}[\sigma(1)] \geq \ldots \geq \boldsymbol{x}[\sigma(n)]$;
  $S_k = \{\sigma(1), \ldots, \sigma(k)\}$;  $K^* = \arg\min_{k \in \{0,1,\ldots,n\}} f(S_k)$;  $C = \{S_k : k \in K^*\}$;
**Output**: Collection of sets $C$, such that $f(A) \leq \tilde{f}(\boldsymbol{x})$ for all $A \in C$

---

### 4.3  Early Stopping based on Discrete Certificates of Optimality

In general, if the minimum of $f$ is not unique, the output of SLG may be in the interior of the unit cube. However, if $f$ admits a unique minimum $A^*$, then the iterates will tend toward the corner $\boldsymbol{e}_{A^*}$. One natural question one may ask, if a trend like this is observed, is it necessary to wait for the iterates to converge all the way to the optimal solution of the continuous problem $\min_{\boldsymbol{x} \in [0,1]^n} \tilde{f}(x)$, when one is actually iterested in solving the discrete problem $\min_{A \in 2^E} f(A)$? Below, we show that it is possible to use information about the current iterates to check optimality of a set and terminate the algorithm before the continuous problem has converged.

To prove optimality of a candidate set $A$, we can use a subgradient of $\tilde{f}$ at $\boldsymbol{e}_A$. If $\boldsymbol{g} \in \partial \tilde{f}(\boldsymbol{e}_A)$, then we can compute an optimality gap:

$$f(A) - f^* \leq \max_{\boldsymbol{x} \in [0,1]^n} (\boldsymbol{e}_A - \boldsymbol{x}) \cdot \boldsymbol{g} = \sum_{k \in A} \max(0, \boldsymbol{g}[k](\boldsymbol{e}_A[k] - \boldsymbol{e}_{E \setminus A}[k])). \quad (5)$$

In particular if $\boldsymbol{g}[k] \leq 0$ for $k \in A$ and $\boldsymbol{g}[k] \geq 0$ for $k \in E \setminus A$, then $A$ is optimal. But if we only have knowledge of candidate set $A$, then finding a subgradient $\boldsymbol{g} \in \partial \tilde{f}(\boldsymbol{e}_A)$ which demonstrates optimality may be extremely difficult, as the set of subgradients is a polyhedron with exponentially many extreme points. But our algorithm naturally suggests the subgradient we could use; the gradient of the smoothed extension is one such subgradient – provided a certain condition is satisfied, as described in the following Lemma.

**Lemma 1** *Suppose $f$ is a decomposable submodular function, with Lovász extension $\tilde{f}$, and smoothed extension $\tilde{f}^\mu$ as in the previous section. Suppose $\boldsymbol{x} \in \mathbb{R}^n$ and $A \in 2^E$ satisfy the following property:*

$$\min_{k \in A, l \in E \setminus A} \boldsymbol{x}[k] - \boldsymbol{x}[l] \geq 2\mu$$

*Then $\nabla \tilde{f}^\mu(\boldsymbol{x}) \in \partial \tilde{f}(\boldsymbol{e}_A)$*

This is a consequence of our formula for $\nabla \tilde{\Psi}^\mu$, but see the appendix for a detailed proof. Lemma 1 states that if the components of point $\boldsymbol{x}$ corresponding to elements of $A$ are all larger than all the other components by at least $2\mu$, then the gradient at $\boldsymbol{x}$ is a subgradient for $\tilde{f}$ at $\boldsymbol{e}_A$ (which by Equation 5 allows us to compute an optimality gap). In practice, this separation of components naturally occurs as the iterates move in the direction of the point $\boldsymbol{e}_A$, long before they ever actually reach the point $\boldsymbol{e}_A$. But even if the components are not separated, we can easily add a positive multiple of $\boldsymbol{e}_A$ to separate them and then compute the gradient there to get an optimality gap. In summary, we have the following algorithm to check the optimality of a candidate set:    Of critical

---

**Algorithm 3:** Set Optimality Check

**Input**: Set $A$; decomposable function $f$; scale $\mu$; $\boldsymbol{x} \in \mathbb{R}^n$.
**begin**
  $\gamma = 2\mu + \max_{k \in A, l \in E \setminus A} \boldsymbol{x}[l] - \boldsymbol{x}[k]$;  $\boldsymbol{g} = \nabla \tilde{f}^\mu(\boldsymbol{x} + \gamma \boldsymbol{e}_A)$;
  $gap = \sum_{k \in A} \max(0, \boldsymbol{g}[k](\boldsymbol{e}_A[k] - \boldsymbol{e}_{E \setminus A}[k]))$;
**Output**: $gap$, which satisfies $gap \geq f(A) - f^*$

---

importance is how to choose the candidate set $A$. But by Equation 5, for a set to be optimal, we want the components of the gradient $\nabla \tilde{f}^\mu(\boldsymbol{A} + \gamma \boldsymbol{e}_A)[k]$ to be negative for $k \in A$ and positive for $k \in E \setminus A$. So it is natural to choose $A = \{k : \nabla \tilde{f}^\mu(\boldsymbol{x})[k] \leq 0\}$. Thus, if adding $\gamma \boldsymbol{e}_A$ does not change the signs of the components of the gradient, then in fact we have found the optimal set. This stopping criterion is very effective in practice, and we use it in all of our experiments.

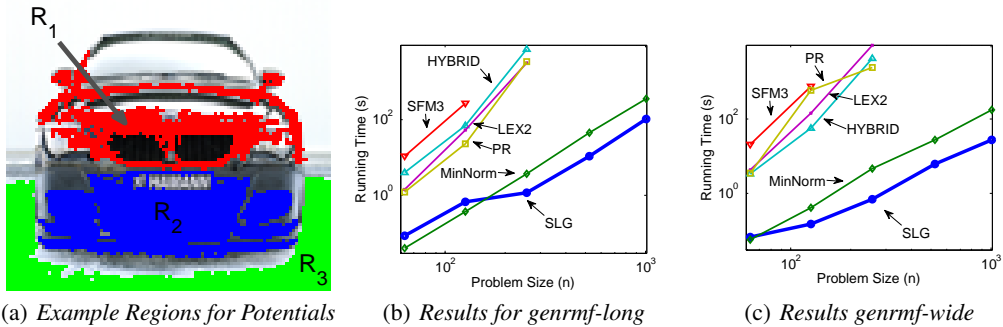

(a) *Example Regions for Potentials*  (b) *Results for genrmf-long*  (c) *Results genrmf-wide*

Figure 1: (a) Example regions used for our higher-order potential functions (b-c) Comparision of running times of submodular minimization algorithms on synthetic problems from DIMACS [1].

## 5 Extension to General Concave Potentials

To extend our algorithm to work on general concave functions, we note that an arbitrary concave function can be expressed as an integral of threshold potential functions. This is a simple consequence of integration by parts, which we state in the following lemma:

**Lemma 2** *For $\phi \in C^2([0,T])$*,

$$\phi(x) = \phi(0) + \phi'(T)x - \int_0^T \min(x,y)\phi''(y)dy, \quad \forall x \in [0,T]$$

This means that for a general sum of concave potentials as in Equation (3), we have:

$$f(A) = \boldsymbol{c} \cdot \boldsymbol{e}_A + \sum_j \left( \phi_j(0) + \phi'(\boldsymbol{w}_j \cdot \mathbf{1})\boldsymbol{w}_j \cdot \boldsymbol{e}_A - \int_0^{\boldsymbol{w}_j \cdot \mathbf{1}} \Psi_{\boldsymbol{w}_j, y}(A)\phi_j''(y)dy \right).$$

Then we can define $\tilde{f}$ and $\tilde{f}^\mu$ by replacing $\Psi$ with $\tilde{\Psi}$ and $\tilde{\Psi}^\mu$ respectively. Our SLG algorithm is essentially unchanged, the conditions for optimality still hold, and so on. Conceptually, we just use a different smoothed gradient, but calculating it is more involved. We need to compute the integrals of the form $\int \nabla \tilde{\Psi}^\mu_{\boldsymbol{w}, y}(\boldsymbol{x})\phi''(y)dy$. Since $\nabla \tilde{\Psi}^\mu_{\boldsymbol{w}, y}(\boldsymbol{x})$ is a piecewise linear function with repect to $y$ which we can compute, we can evaluate the integral by parts so that we need only evaluate $\phi$, but not its derivatives. We leave this formula for the appendix.

## 6 Experiments

**Synthetic Data.** We reproduce the experimental setup of [8] designed to compare submodular minimization algorithms. Our goal is to find the minimum cut of a randomly generated graph (which requires submodular minimization of a sum of 2-potentials) with the graph generated by the specifications in [1]. We compare against the state of the art combinatorial algorithms (LEX2, HYBRID, SFM3, PR [6]) that are guaranteed to find the exact solution in polynomial time, as well as the Minimum Norm algorithm of [8], a practical alternative with unknown running time. Figures 1(b) and 1(c) compare the running time of SLG against the running times reported in [8]. In some cases, SLG was 6 times faster than the MinNorm algorithm. However the comparison to the MinNorm algorithm is inconclusive in this experiment, since while we used a faster machine, we also used a simple MATLAB implementation. What is clear is that SLG scales at least as well as MinNorm on these problems, and is practical for problem sizes that the combinatorial algorithms cannot handle.

**Image Segmentation Experiments.** We also tested our algorithm on the joint image segmentation-and-classification task introduced in Section 3. We used an implementation of TextonBoost [20], then trained on and tested subsampled images from [5]. As seen in Figures 2(e) and 2(g), using only the per-pixel score from our TextonBoost implementation gets the general area of the object, but does not do a good job of identifying the shape of a classified object. Compare to the ground truth in Figures 2(b) and 2(d). We then perform MAP inference in a Markov Random Field with 2-potentials (as done in [20]). While this regularization, as shown in Figures 2(f) and 2(h), leads to improved performance, it still performs poorly on classifying the boundary.

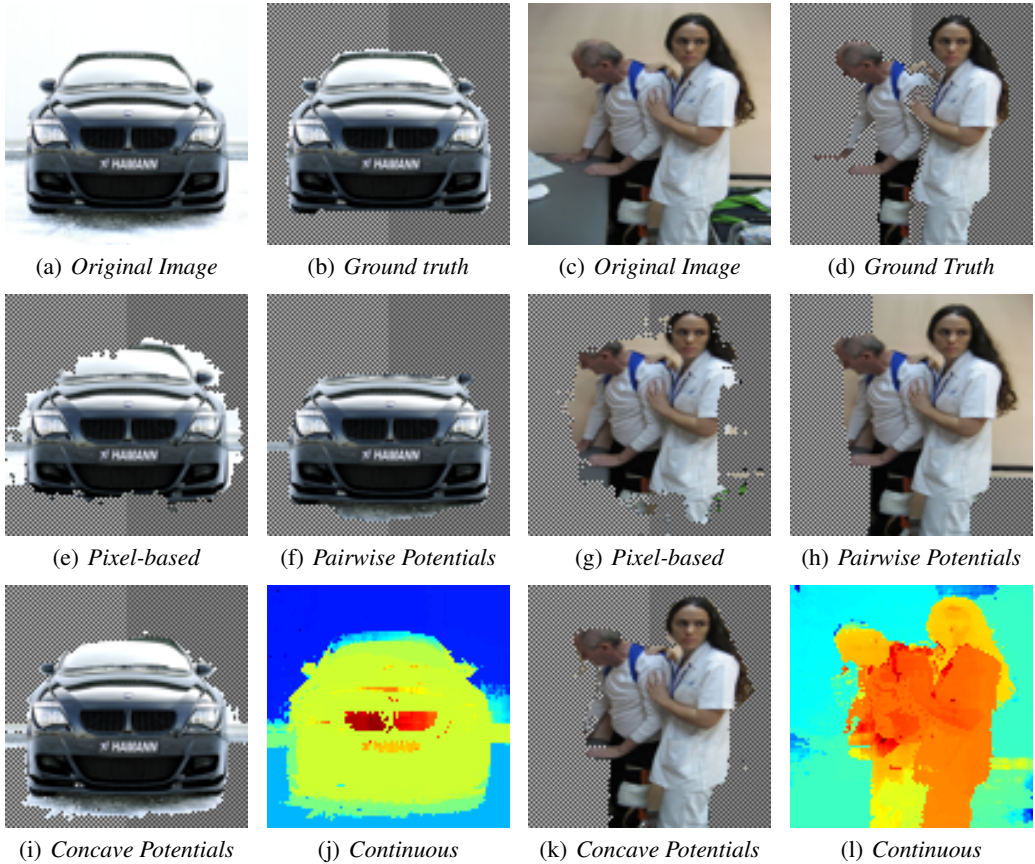

Figure 2: Segmentation experimental results

Finally, we used SLG to regularize with higher order potentials. To generate regions for our potentials, we randomly picked seed pixels and grew the regions based on HSV channels of the image. We picked our seed pixels with a preference for pixels which were included in the least number of previously generated regions. Figure 1(a) shows what the regions typically looked like. For our experiments, we used 90 total regions. We used SLG to minimize $f(A) = \boldsymbol{c} \cdot \boldsymbol{e}_A + \sum_j |A \cap R_j||R_j \setminus A|$, where $\boldsymbol{c}$ was the output from TextonBoost, scaled appropriately. Figures 2(i) and 2(k) show the classification output. The continuous variables $\boldsymbol{x}$ at the end of each run are shown in Figures 2(j) and 2(l); while it has no formal meaning, in general one can interpret a very high or low value of $x[k]$ to correspond to high confidence in the classification of the pixel $k$. To generate the result shown in Figure 2(k), a problem with $10^4$ variables and 90 concave potentials, our MATLAB/mex implementation of SLG took 71.4 seconds. In comparison, the MinNorm implementation of the SFO toolbox [15] gave the same result, but took 6900 seconds. Similar problems on an image of twice the resolution ($4 \times 10^4$ variables) were tested using SLG, resulting in a runtimes of roughly 1600 seconds.

## 7 Conclusion

We have developed a novel method for efficiently minimizing a large class of submodular functions of practical importance. We do so by decomposing the function into a sum of threshold potentials, whose Lovász extensions are convenient for using modern smoothing techniques of convex optimization. This allows us to solve submodular minimization problems with thousands of variables, that cannot be expressed using only pairwise potentials. Thus we have achieved a middle ground between graph-cut-based algorithms which are extremely fast but only able to handle very specific types of submodular minimization problems, and combinatorial algorithms which assume nothing but submodularity but are impractical for large-scale problems.

**Acknowledgements**   This research was partially supported by NSF grant IIS-0953413, a gift from Microsoft Corporation and an Okawa Foundation Research Grant. Thanks to Alex Gittens and Michael McCoy for use of their TextonBoost implementation.

## A  Submodularity of Decomposable Functions

Since the sum of submodular functions is submodular, we need only prove that the submodularity of $f(A) = \phi(\boldsymbol{w} \cdot \boldsymbol{e}_A)$, where $\phi$ is an arbitrary concave function on $\mathbb{R}$ and $\boldsymbol{w} \geq \boldsymbol{0}$.

By definition of concavity, for all $\theta \in [0, 1]$, we have:
$$\phi(\theta(y + h) + (1 - \theta)x) + \phi((1 - \theta)(y + h) + \theta x) \geq \phi(y + h) + \phi(x)$$
If $x \leq y$ and $h \geq 0$, then setting $\theta = h/(y - x + h)$ in the above gives us:
$$\phi(x + h) - \phi(x) \geq \phi(y + h) - \phi(y) \tag{6}$$
Then, for all $k \in E \setminus A$, we compute the the discrete derivative $\Delta_k f(A)$:
$$\Delta_k f(A) = \phi(\boldsymbol{w} \cdot \boldsymbol{e}_A + \boldsymbol{w}[k]) - \phi(\boldsymbol{w} \cdot \boldsymbol{e}_A) \tag{7}$$
So if $A \subseteq B \subseteq E$ and $k \in E \setminus B$, then $\boldsymbol{w} \cdot \boldsymbol{e}_A \leq \boldsymbol{w} \cdot \boldsymbol{e}_B$, so by Eqs. 6 and 7, $\Delta_k f(A) \geq \Delta_k f(B)$, and hence $f$ is submodular.

## B  Reformulation of Set Cover Functions

A set cover function can be formulated as the function:
$$f(A) = |\cup_{i \in A} B_i|$$
Where $B_i$ are subsets of some base set $F$, and the $B_i$ form some collection of subsets indexed by $E$. For every $k \in F$, we define the vectors $\boldsymbol{w}_k \in R^{|E|}$ as follows:
$$\boldsymbol{w}_k[i] = \begin{cases} 0 & k \notin B_i \\ 1 & k \in B_i \end{cases}$$
We claim:
$$f(A) = \sum_{k \in F} \min(1, \boldsymbol{w}_k \cdot \boldsymbol{e}_A)$$
The $k$th term in the sum equals 1 if $k \in B_i$ for some $i \in A$ and 0 otherwise. The sum of all such terms will give the cardinality of the union of the $B_i$ with $i \in A$, which is exactly the set cover function.

## C  Strict Generality of Threshold Potentials

As mentioned in the text, any concave cardinality function can be decomposed into the sum of several threshold potentials. This is effectively the discrete version of Lemma 2:
$$\phi(|A \cap S|) = \phi(0) + (\phi(|S|) - \phi(|S| - 1))\boldsymbol{e}_S \cdot \boldsymbol{e}_A + \sum_{k=1}^{|S|-1} (2\phi(k) - \phi(k-1) - \phi(k+1)) \min(k, \boldsymbol{e}_S \cdot \boldsymbol{e}_A)$$
Since $\phi$ is concave, the coefficients $(2\phi(k) - \phi(k-1) - \phi(k+1))$ are nonnegative. So without loss of generality, any sum of concave cardinality functions can be expressed as a sum of a modular function and nonnegative linear combination of threshold potentials:
$$\sum_j \phi_j(|R_j \cap A|) = \boldsymbol{c} \cdot e_A + \sum_{S_k \subset E, |S|>1} \sum_{k=1}^{|S|-1} d_{kl} \min(k, \boldsymbol{e}_{S_k} \cdot \boldsymbol{e}_A)$$
There are $\sum_{m=2}^{n} \binom{n}{m}(m-1)$ coefficients $d_{kl}$ and they all must be nonnegative. So to check if a submodular function $f(A)$ can be expressed as such a sum, we can just write out the $2^n$ constraints for each subset:
$$f(A) = \boldsymbol{c} \cdot e_A + \sum_{S_k \subset E, |S|>1} \sum_{k=1}^{|S|-1} d_{kl} \min(k, \boldsymbol{e}_{S_k} \cdot \boldsymbol{e}_A) \quad \text{for all } A \in 2^E \tag{8}$$
If $n = 4$, we have $2^4$ linear constraints, 4 unconstrained variables from $\boldsymbol{c}$, and 19 nonnegative variables $d_{kl}$. This is small enough that one can check for feasibility using a linear algebra package. We discovered that simple threshold potential $f(A) = \min(y, \boldsymbol{w} \cdot \boldsymbol{e}_A)$ with $\boldsymbol{w} = [1, 2, 3, 4]/4$ and $y = 1$ does not have a feasible solution to Eq. 8.

## D  Reformulation of a Class of Functions

Another example of decomposable functions are the problems under consideration in [10], which are of the following form:
$$f(A) = \boldsymbol{c} \cdot \boldsymbol{e}_A + (\boldsymbol{u} \cdot \boldsymbol{e}_A)\phi(\boldsymbol{v} \cdot \boldsymbol{e}_A)$$

Where $\boldsymbol{u}, \boldsymbol{v}$ are nonnegative weight vectors and $\phi$ is a nonincreasing concave function. Suppose we can choose vectors $\boldsymbol{w}_j$ and concave $\tilde{\phi}$ to satisfy:

$$\tilde{\phi}(\boldsymbol{w}_j \cdot \boldsymbol{e}_A) = \begin{cases} 0 & \text{if } j \notin A \\ \phi(\boldsymbol{v} \cdot \boldsymbol{e}_A) - \phi(0) & \text{if } j \in A \end{cases} \tag{9}$$

Then we claim the following is an equivalent formulation for $f$ in decomposable form:

$$f'(A) = (\boldsymbol{c} + \phi(0)\boldsymbol{u}) \cdot \boldsymbol{e}_A + \sum_{j=1}^{n} \boldsymbol{u}[j]\tilde{\phi}(\boldsymbol{w}_j \cdot \boldsymbol{e}_A) \tag{10}$$

Indeed, plugging Eq. 9 to the above gives:

$$f'(A) = (\boldsymbol{c} + \phi(0)\boldsymbol{u}) \cdot \boldsymbol{e}_A + \sum_{j \in A} \boldsymbol{u}[j](\phi(\boldsymbol{v} \cdot \boldsymbol{e}_A) - \phi(0)) = f(A)$$

To satisfy Eq. 9 we define $\tilde{\phi}$ as follows:

$$\tilde{\phi}(t) = \begin{cases} 0 & \text{if } t \leq \mathbf{1} \cdot \boldsymbol{v} \\ \phi(t - \mathbf{1} \cdot \boldsymbol{v}) - \phi(0) & \text{if } t > \mathbf{1} \cdot \boldsymbol{v} \end{cases}$$

And let $\boldsymbol{w}_j = \boldsymbol{v} + (\mathbf{1} \cdot \boldsymbol{v})\boldsymbol{e}_j$. It is straightforward to check that these definitions satisfy Eq. 9. Note $\tilde{\phi}$ is concave because $\phi$ is nonincreasing concave. Incidentally, the decomposition in Eq. 10 proves that $f$ is submodular.

## E  Proof of Lemma 1

By linearity, it is sufficient to consider the case $f = \Psi^{\mu}_{\boldsymbol{w},y}$. First we claim that if the hypothesis of the Lemma holds, adding a positive multiple of $\boldsymbol{e}_A$ will not change the gradient. That is, $\nabla \tilde{\Psi}^{\mu}_{\boldsymbol{w},y}(\boldsymbol{x}) = \nabla \tilde{\Psi}^{\mu}_{\boldsymbol{w},y}(\boldsymbol{x} + \alpha \boldsymbol{e}_A)$ for $\alpha > 0$. Recall the formula for the gradient:

$$\nabla \tilde{\Psi}^{\mu}_{\boldsymbol{w},y}(\boldsymbol{x}) = \min(\max((\boldsymbol{x} - t^*\mathbf{1})/\mu, \mathbf{0})$$
$$\text{where } t^* \text{ satisfies } \min(\max((\boldsymbol{x} - t^*\mathbf{1})/\mu, \mathbf{0}), \boldsymbol{w}) \cdot \mathbf{1} = y$$

Consider the effect of adding $\alpha \boldsymbol{e}_A$ to $\boldsymbol{x}$ in this formula; either $t^*$ is increased by $\alpha$ or it is unchanged; in either case the gradient itself is unchanged. Next, note the following scale relationship which follows directly from the definition of $\tilde{\Psi}^{\mu}_{\boldsymbol{w},y}$:

$$\tilde{\Psi}^{\mu}_{\boldsymbol{w},y}(\alpha \boldsymbol{x}) = \alpha \tilde{\Psi}^{\mu/\alpha}_{\boldsymbol{w},y}(\boldsymbol{x}).$$

But combined with our first observation this implies

$$\nabla \tilde{\Psi}^{\mu}_{\boldsymbol{w},y}(\boldsymbol{x}) = \nabla \tilde{\Psi}^{\mu/\alpha}_{\boldsymbol{w},y}(\boldsymbol{x}/\alpha + \boldsymbol{e}_A)$$

But the right-hand side of that equation must converge to a subgradient of the nonsmooth function as $\alpha \to \infty$:

$$\lim_{\alpha \to \infty} \nabla \tilde{\Psi}^{\mu/\alpha}_{\boldsymbol{w},y}(\boldsymbol{x}/\alpha + \boldsymbol{e}_A) \in \partial \tilde{\Psi}_{\boldsymbol{w},y}(\boldsymbol{e}_A)$$

which gives the result.

# F Proof of Lemma 2

This is straightforward calculation:

$$
\begin{aligned}
\int_0^T \min(x,y)\phi''(y)dy &= \int_0^x y\phi''(y)dy + \int_x^T x\phi''(y)dy \\
&= (y\phi'(y) - \phi(y))\big|_0^x + x\phi'(y)\big|_x^T \\
&= x\phi'(x) - \phi(x) + \phi(0) + x\phi'(T) - x\phi'(x) \\
&= \phi(0) + x\phi'(T) - \phi(x)
\end{aligned}
$$

Intuitively, this is a consequence of integration by parts and the fact that $\frac{\partial^2}{\partial x^2}\min(x,y) = -\delta(x-y)$ (the Dirac delta).

# G General Smoothed Gradient Formula

Let $f(A) = \phi(\boldsymbol{w} \cdot \boldsymbol{e}_A)$ be a general concave potential. For ease of notation, in the following let $\boldsymbol{g}(y) = \nabla \tilde{\Psi}^\mu_{\boldsymbol{w},y}(\boldsymbol{x})$ be the gradient of the smoothed extension of a threshold potential. Then by Lemma 2, we have this formula for the gradient of smoothed extention of $f$:

$$\nabla \tilde{f}^\mu(\boldsymbol{x}) = \phi'(\boldsymbol{w} \cdot \boldsymbol{1})\boldsymbol{w} - \int_0^{\boldsymbol{w} \cdot \boldsymbol{1}} \boldsymbol{g}(y)\phi''(y)dy$$

Note that $\boldsymbol{g}$ is a piecewise linear function of $y$. Let the intervals $[y_i, y_{i+1}]$ with $0 = y_0 \leq \ldots \leq y_N = \boldsymbol{w} \cdot \boldsymbol{1}$ be the intervals that $\boldsymbol{g}$ is linear on. Let $\boldsymbol{\theta}_i = \boldsymbol{g}(y_i)$, so then $\boldsymbol{0} = \boldsymbol{\theta}_0 \leq \ldots \leq \boldsymbol{\theta}_N = \boldsymbol{w}$. Finally let $\boldsymbol{g}_i(y)$ be the linear functions that $\boldsymbol{g}$ equals on these intervals,:

$$\boldsymbol{g}(y) = \boldsymbol{g}_i(y) \text{ for } y \in [y_{i-1}, y_i]$$

Denote by $\boldsymbol{g}'_i = (\boldsymbol{\theta}_i - \boldsymbol{\theta}_{i-1})/(y_i - y_{i-1})$ the vector which is derivative of $\boldsymbol{g}_i(y)$ with respect to $y$. So then our smoothed gradient can be evaluated:

$$
\begin{aligned}
\nabla \tilde{f}^\mu(\boldsymbol{x}) &= \phi'(y_N)\boldsymbol{w} - \sum_{i=1}^N \int_{y_{i-1}}^{y_i} \boldsymbol{g}_i(y)\phi''(y)dy \\
&= \phi'(y_N)\boldsymbol{w} + \sum_{i=1}^N (\boldsymbol{g}'_i\phi(y) - \boldsymbol{g}_i(y)\phi'(y))\big|_{y_{i-1}}^{y_i} \\
&= \phi'(y_N)\boldsymbol{w} + \sum_{i=1}^N \boldsymbol{g}'_i(\phi(y_i) - \phi(y_{i-1})) - \sum_{i=1}^N (\boldsymbol{\theta}_i\phi'(y_i) - \boldsymbol{\theta}_{i-1}\phi'(y_{i-1})) \\
&= \sum_{i=1}^N \frac{(\boldsymbol{g}(y_i) - \boldsymbol{g}(y_{i-1}))(\phi(y_i) - \phi(y_{i-1}))}{y_i - y_{i-1}}
\end{aligned}
$$

Note there are at most $2n$ points $y_i$, and they can be found all in $O(n \log n)$ time, since it requires a sort. So the overall operation count of evaluating this formula is $O(n^2)$ since it requires adding up $O(n)$ $n$-dimensional vectors.